\documentclass[11pt]{article}
\usepackage{acl}

\usepackage{times}
\usepackage{latexsym}
\usepackage[T1]{fontenc}
\usepackage[utf8]{inputenc}
\usepackage{microtype}
\usepackage{amsmath}
\usepackage{amssymb}
\usepackage{booktabs}
\usepackage{graphicx}
\usepackage{inconsolata}
\usepackage{enumitem}
\setlist{noitemsep,topsep=2pt}
\usepackage{listings}
\title{Fallacy Benchmarks Measure Scheme Recognition,\\Not Fallacy Detection}

\author{Navyansh Singh \\
  IIIT Naya Raipur \\
  {\small\texttt{navyansh24102@iiitnr.edu.in}} \And
  Animesh Pathak \\
  IIIT Naya Raipur \\
  {\small\texttt{animesh24100@iiitnr.edu.in}} \And
  Aarav Singh \\
  IIIT Naya Raipur \\
  {\small\texttt{aarav24101@iiitnr.edu.in}}}

\begin{document}
\maketitle

\begin{abstract}
Fallacy-detection benchmarks pair fallacy classes with a single ``valid'' or ``none'' class that takes everything data collection did not label as a fallacy. A detector has two jobs, deciding whether an argument is fallacious and naming which fallacy it commits, and the false-positive rate is meant to measure the first. We show that what these benchmarks actually score is scheme recognition, the ability behind the second job. Their own test sets already show it: when a classifier misses a fallacy, the error lands on ``none'' rather than on another fallacy type, so detection is failing while classification holds. The reason is what the valid class lacks. The negatives that separate the two jobs are correct arguments using the same argumentation scheme as a fallacy, and they are scarce: nearly absent from the four benchmarks we examined, and rare even under deliberate search. A detector is therefore never tested where recognizing a scheme and judging its use come apart, and can pass on recognition alone. We construct the missing arguments, together with a control condition from the same pipeline that differs only in scheme, so whatever generation contributes, it contributes to both. The classifier labels the scheme-matched negatives as the source fallacy, and labels the wrong-scheme negatives as the scheme they actually use 85.9\% of the time and as the source type 0.4\%. The classifier has learned which scheme an argument uses, not whether it uses it correctly. The over-flagging follows: a model that scores 16.6\% on CoCoLoFa's own valid class flags 58.9\% of the constructed arguments. The same dissociation appears in three zero-shot LLM detectors that never saw these benchmarks. We release the items as Scheme Foils. A reported false-positive rate should not be trusted as a measure of detection until the valid class has been audited for scheme-matched coverage.
\end{abstract}

\section{Introduction}
\label{sec:intro}

Fallacy-detection benchmarks are classification datasets. A span of text is assigned to one of several fallacy types, or to a ``valid'' or ``none'' class holding everything not judged fallacious \citep{yeh2024cocolofa,helwe2024mafalda,habernal2017argotario}. A fallacy detector has two jobs: \emph{detection}, separating fallacious arguments from valid ones, and \emph{classification}, naming the type of fallacy an argument commits. What decides whether a detector is usable is detection---flagging the arguments that contain a fallacy and passing the ones that do not---and its measure on sound reasoning is the false-positive rate. On standard benchmarks that rate is small, and is reported as evidence that over-flagging is minor \citep{yeh2024cocolofa,singh2026debiasing}.

We show that the number is small because of how the valid class is assembled. It is not built the way the fallacy classes are built. It is the residue: whatever crowd workers produced when asked \emph{not} to write a fallacy, whatever a forum scrape left over, whatever an annotation protocol swept into ``none.'' We sampled the valid class of four benchmarks and found almost none of it scheme-matched---using the same reasoning pattern as a fallacy without committing it. The classes hold off-topic remarks, bare opinions, fragments, comments \emph{about} fallacies, and genuine arguments, most built on reasoning patterns no fallacy class corresponds to. Separating that material from a fallacy takes no ability to tell valid reasoning from fallacious reasoning---only recognition of which pattern an argument uses. What is missing is the scheme-matched valid argument, and two independent annotators found 3 such items in 183 sampled.

That population is the one detection should be measured on. A valid appeal to expert opinion sits close to a fallacious one; what separates them is whether the cited authority is genuinely qualified, not the argument's surface shape. Walton's argumentation schemes \citep{walton2008schemes} make this operational (\S\ref{sec:background}) and give each fallacy type its matched negative: a valid argument answering the question the fallacy leaves open.

To measure detection on the population where the two abilities come apart, we construct matched negatives for eight fallacy types across two corpora. A fine-tuned classifier that flags 16.6\% of CoCoLoFa's native valid class flags 58.9\% of them; trained from scratch on Reddit, 5.7\% becomes 62.0\%.

A rate measured on constructed items depends on how the items were written, so for each source item the same pipeline also produces a \emph{wrong-scheme} negative: an equally careful valid argument on the same topic, at the same length and register, built on a different scheme. The conditions differ in scheme identity and nothing else---any signature of machine generation is present in both and cannot produce a difference, and ``carefully written is simply harder'' predicts no difference either. We compare them on how often each is classified as the \emph{source} type specifically; that comparison is invariant to generation quality, which no raw rate on constructed data can be.

The two conditions come apart sharply. Matched negatives are classified as their source type 40.9 points more often than wrong-scheme negatives; wrong-scheme negatives are instead classified as the scheme they do use 85.9\% of the time, and as the source type 0.4\% of the time. The raw rates establish that a gap exists; the dissociation establishes what it is made of.

The dissociation is robust. Annotators blind to classifier behaviour confirm the matched negatives valid, and the gap survives on the confirmed subset. It grows when the domain gap is removed by training from scratch---the opposite of what domain shift predicts. It appears in three zero-shot detectors that never saw the benchmark. And the over-flagging does not need our pipeline at all: on the unmodified benchmark the classifier's own errors on fallacies land on ``none'' rather than on another type; it appears on naturally occurring arguments no one wrote for this study; and, attenuated, on a negative class an independent shared task built deliberately around this population.

Detectors track scheme identity and not critical-question failure, and a valid class containing no correctly-instantiated schemes contains no item on which those two come apart. The benchmark therefore scores scheme recognition---the classification-shaped skill---and reports it as detection. Its false-positive rate is low for a reason that has nothing to do with the detector being right.

\section{Scheme-Matched Negatives and the Valid Class}
\label{sec:background}

\paragraph{Schemes and critical questions.} We take our argumentation schemes from Walton \citep{walton1996schemes,walton2008schemes}. A scheme is a reusable pattern of everyday inference, and each scheme carries critical questions: the checks an argument of that pattern must survive. An argument that answers its scheme's critical questions is a valid instance; one that fails a specific question, in the characteristic way, is a \emph{fallacy of type $\tau$}. For expert opinion the questions are whether the person cited is a genuine expert, in the relevant domain, speaking within their competence. A climate argument citing a named climatologist answers them and is valid. The same argument citing a talk-show host fails the first question---the fallacy called ``appeal to authority.'' A negative example \emph{matched} to $\tau$ is exactly this valid counterpart: an argument on the same scheme that answers the question the fallacy leaves open, comparable in topic, length, and register. Throughout, \emph{valid} means what the benchmarks mean by it (the ``valid'' or ``none'' class), not deductive validity; we call that class the benchmark's \emph{native} class when contrasting it with constructed negatives. We use Walton instrumentally, in the operational form of \citet{ruizdolz2023fallacies}.

A classifier that flags a matched negative is tracking the scheme's surface signature, not the critical-question failure its label denotes. The two are separable only on items where a scheme is instantiated correctly, so a valid class without them leaves the substitution invisible. Walton's account predicts these pairs are hard: a fallacy characteristically appears a better argument of its kind than it is, because it borrows the form of a legitimate scheme and fails only at the step a surface reading does not expose \citep{walton2010fallacies}.

Classification is the job a benchmark can score directly, and classification is scheme recognition: telling appeal to authority from slippery slope requires no judgment of whether either is used correctly. A detector optimized for it becomes a scheme recognizer, and nothing in a valid class without correctly-instantiated schemes ever penalizes that.

\paragraph{Why the valid class fills up with easy material.} Each fallacy type is tied to one scheme and one critical-question failure, but validity is tied to no scheme in particular: every correct instantiation of every scheme is valid, and Walton's catalogue alone enumerates more than sixty. ``Valid'' is therefore a far broader category than any single fallacy type, and a class assembled without scheme-level targeting has an abundant easy region to default into. That region is not hypothetical: \citet{feng2011scheme} classify Walton schemes from surface features, and \citet{lawrence2016scheme} independently recover scheme identity from real text. A false-positive rate reported over such a class is a weighted average dominated by the easy majority, so building the valid class without scheme-level targeting does not remove the asymmetry between valid and fallacious arguments. It removes it from the measurement.

How often a deployed detector meets such an argument is beside the point: the reported rate predicts nothing about what happens when it does, and a system built to flag fallacies in argumentation cannot avoid correct arguments.

\section{Related Work}
\label{sec:related}

\paragraph{Fallacy benchmarks.} Nearly all fallacy-detection benchmarks share one shape: fallacy classes plus a single undifferentiated negative class. CoCoLoFa \citep{yeh2024cocolofa}, MAFALDA \citep{helwe2024mafalda}, Argotario \citep{habernal2017argotario} and \citet{sahai2021breaking} follow it; others omit the negative class altogether \citep{jin2022logical,goffredo2022fallacious}, and a benchmark with no valid class cannot measure over-flagging at all. None reports how deliberately its negative class was constructed relative to the schemes its fallacy types instantiate. CoCoLoFa's authors document the gap themselves. Some comment writers, asked to produce a fallacy, instead produced valid scheme instances that experts judged ``fallacy-like but valid.'' The authors note the critical-question-based validity check \citet{ruizdolz2023fallacies} had suggested, and judge forgoing it a reasonable trade-off given its cost and the low rate they observed, 12 of 237 comments \citep{yeh2024cocolofa}.

\paragraph{Dataset artifacts.} A benchmark can report high scores for reasons that lie in how its data was collected rather than in what models understand: \citet{gururangan2018artifacts} recover NLI labels from the hypothesis alone, \citet{mccoy2019right} show inference models applying syntactic heuristics, and \citet{niven2019probing} account for a model's argument-comprehension score entirely by spurious cues. In each case the contribution was the diagnosis and the diagnostic set; none corrected the benchmark it was about. Those artifacts live in the positive class or in the association between the two; ours lives in the composition of the negative class. In measurement terms \citep{jacobs2021measurement}, the operationalization---score over the shipped valid class---no longer measures the conceptualization the label names.

\paragraph{Hard negatives for evaluation.} \citet{gardner2020contrast} show that naturally collected test sets overstate competence and that deliberately constructed hard examples expose the gap, but supply no criterion for which negatives are hard; for argumentation that criterion is not read off the surface, and Walton's apparatus supplies it. In the fallacy domain, \citet{singh2026debiasing} builds counterfactual negatives for fallacy detection and finds the resulting bias real and persistent, without reference to argumentation schemes and without an account of why standard benchmarks fail to surface it. Minimal editing \citep{kaushik2020learning} is the alternative construction we set aside for this measurement, not in general (\S\ref{sec:construction}).

\paragraph{Closest prior work.} \citet{ruizdolz2023fallacies} test already-trained classifiers on fourteen hand-built arguments---one valid and one fallacious instance of each of seven schemes---and find that fine-tuned and generative models alike struggle to separate them. They attribute specific failures to surface-vocabulary matching, the same mechanism we formalize. Their account treats the sharing as a per-instance modelling deficiency to be corrected architecturally; we show the benchmark's aggregate reported rate is deflated by how the valid class is built. Their remedy \citep{ruizdolz2025nlascq} replaces the sequence classifier with a two-stage pipeline requiring annotated data and two models at inference; ours changes no architecture and corrects what the negative class contains.

\paragraph{Concurrent work.} The Touch\'e 2026 shared task on fallacy detection \citep{touche2026} builds its non-fallacious class by selecting arguments whose reasoning pattern resembles a specific fallacy type, so that systems must separate genuinely fallacious arguments from superficially similar valid ones. Their negatives are obtained by selection: items are not counterparts of particular fallacious sources, so topic, length and register are not held fixed and no wrong-scheme comparison is available. Their scheme dimension follows Macagno's goal/basis axes \citep{macagno2022profiles}, so resemblance is judged at selection time, not by which critical question an argument answers. We audit their negative class in \S\ref{sec:touche}.

\section{Constructing Scheme-Matched Negatives}
\label{sec:construction}

\paragraph{Fresh writing, not minimal editing.} We do not build negatives by minimal editing \citep{kaushik2020learning}. A minimal edit turns a fallacious item into a valid one while keeping most of its wording, so it holds two things fixed at once: the lexical surface and the scheme. A flag on such a negative cannot be attributed cleanly---the classifier may be reading the scheme, or just the leftover words of a known fallacy. And the wrong-scheme condition below could not be built by minimal edits at all: changing the scheme means rewriting the reasoning. We therefore write every negative from scratch---a new valid argument on the same topic, in the same register and length range, using the required scheme without reusing the source's wording. The matched negative then shares its source's scheme, not its sentences.

\paragraph{The wrong-scheme condition.} For each fallacious source item of type $\tau$ we generate two negatives. The matched negative uses $\tau$'s scheme and answers its critical question. The wrong-scheme negative is a valid argument on the same topic, at the same length and register, using a different scheme $\tau'$, assigned per item and balanced across the other seven types under a fixed seed. The two conditions are built identically except for the scheme.

We compare the conditions on one statistic: how often a negative is classified as the source type $\tau$ specifically. Generic difficulty predicts more flags of every kind; only scheme-tracking predicts flags of that one type. Both predictions are fixed before any measurement.

A false-positive rate on generated negatives can be moved: change the prompt, the model or the filter and the number changes. The difference between the two conditions cannot, because all three are shared---whatever the generator contributes, it contributes to both. Hand-built pairs \citep{ruizdolz2023fallacies}, minimal edits \citep{kaushik2020learning,singh2026debiasing} and selected hard negatives all vary the item itself; ours varies only the scheme. Fresh writing removes shared wording as an explanation for a flag; the wrong-scheme control removes topic and generic difficulty; what remains to explain a difference between the conditions is scheme content.

\paragraph{Generation and filtering.} Generation uses Gemini 2.5 Flash with one prompt design per fallacy type, stating the target scheme and its critical question explicitly (prompts in Appendix~\ref{app:prompts}). Automated checks on register, readability and lexical overlap are diagnostic, not drop criteria. No item was removed for register mismatch; when \S\ref{sec:central} finds no independent register effect, that is the data as generated, not the filter. Items are then judged by a cross-model judge, DeepSeek, chosen for a different architecture and training lineage from the generator; position bias is controlled by running each pairwise comparison in both orders. Matched negatives are judged on scheme fidelity, critical-question satisfaction and conclusion preservation; wrong-scheme negatives on register, on-topicality and scheme switching only, so a retained wrong-scheme item is certified scheme-switched and on-topic, not certified a valid instance of $\tau'$ (\S\ref{sec:wsvalidity}). On CoCoLoFa the judge dropped 128 matched items (124 scheme drift, 4 conclusion flips) and 58 wrong-scheme items, leaving 655 and 738; Reddit and Argotario yielded 387/389 and 219/246 under the same pipeline.

\paragraph{The audit.} The procedure generalizes to any benchmark. Steps 1, 2 and 5 require only an existing benchmark, a coding of its negative class, and a classifier already trained on it; steps 3 and 4 are the remedy.

\begin{enumerate}
\item \emph{Map each fallacy type to its scheme and critical question} \citep{walton2008schemes}; without such a target, ``matched negative'' is undefined. For CoCoLoFa's eight types the mapping is direct for five; false dilemma, appeal to nature and appeal to worse problems sit less cleanly in Walton's catalogue.
\item \emph{Audit the negative class for scheme-matched coverage.} Sample it and code each item as a genuine instance of some tracked scheme or not (\S\ref{sec:composition}). If such arguments are rare, the benchmark is not testing the hard case.
\item \emph{Construct matched negatives where coverage is thin}, holding topic, length and register comparable. For the eight types studied here this step can be skipped by scoring against Scheme Foils directly.
\item \emph{Validate construction, not fluency}: check scheme fidelity and conclusion preservation with automated checks plus a judge from a different model lineage than the generator.
\item \emph{Measure two quantities}: the false-positive rate over the native valid class against the rate over matched negatives, and the wrong-scheme control---how often each condition is classified as $\tau$ specifically.
\end{enumerate}

\section{Experimental Setup}
\label{sec:setup}

\paragraph{Corpora.} CoCoLoFa \citep{yeh2024cocolofa} is the primary corpus and the hardest case for our claim, since its ``none'' negatives look most genuinely argumentative on inspection. Reddit comment threads \citep{sahai2021breaking} are the second within-corpus site, and the one that rules out domain shift because we can train on it from scratch. Argotario \citep{habernal2017argotario} and MAFALDA \citep{helwe2024mafalda} enter the negative-class characterization of \S\ref{sec:composition} and provide corroborating checks (Table~\ref{tab:corrob}, Appendix~\ref{app:tables}).

\paragraph{Classifiers.} The primary classifier is ModernBERT-base \citep{warner2025modernbert} fine-tuned on CoCoLoFa's training split as a nine-way classifier over the eight fallacy types plus ``none.'' A single CoCoLoFa-trained model across corpora isolates negative-class construction as the variable of interest. To rule out domain shift we separately train from scratch on Reddit and evaluate within that corpus; to check architecture-independence we fine-tune RoBERTa-base \citep{liu2019roberta} under the same protocol (five seeds; three each for Reddit and RoBERTa). Every checkpoint was selected by held-out dev accuracy, never by any false-positive-related metric: selecting downstream of the central quantity would make the comparison circular. All models sit well above their majority baselines (Appendix~\ref{app:training}).

\section{Results}
\label{sec:results}

Four experiments test the same claim, each on different data. Only one uses constructed data: the matched/wrong-scheme comparison (\S\ref{sec:central}, repeated across detectors in \S\ref{sec:llm}). The others use where the unmodified benchmark's errors fall (\S\ref{sec:errors}), what the native valid classes contain (\S\ref{sec:composition}), and arguments no one wrote for this study (\S\ref{sec:natural}). The audit also runs on a negative class someone else built (\S\ref{sec:touche}), and two studies validate the constructed items (\S\ref{sec:human}, \S\ref{sec:wsvalidity}).

\subsection{Detection fails on the benchmark's own test set}
\label{sec:errors}

The benchmark's own errors already show which job is failing. When the fine-tuned ModernBERT classifier misclassifies a fallacious CoCoLoFa test item, it does not usually pick the wrong fallacy: 76.9\% of those errors land on ``none'' (95\% CI [64.8\%, 86.7\%]). Had classification been the failing job, the errors would spread across the other seven types; the model separates those types well, and they do not. A TF-IDF baseline sends 97.1\% of its errors to ``none,'' a weak-classifier floor, and a permutation test preserving each model's rate of predicting ``none'' gives $p < 0.001$ for both. Nothing here is constructed, generated or annotated by us: this is the shipped model on the shipped test set. On the benchmark as published, detection is failing while classification holds.

\subsection{The valid class is mostly unmatched material}
\label{sec:composition}

Two independent annotators coded 183 native ``none'' items---samples of 30, 30 and 60 from CoCoLoFa, Reddit and Argotario, and all 63 items of MAFALDA's extractable none class---as a scheme-matched valid argument (A), a genuine argument instantiating no tracked scheme (B), or not an argument (Table~\ref{tab:composition}). They agreed on 86.3\% of items (three-way $\kappa = 0.74$, Gwet's AC1 0.82; positive specific agreement on A, 0.50) and resolved disagreements by discussion without the authors (protocol in Appendix~B). Category A is 1 of 30 CoCoLoFa items, 1 of 30 Reddit, 1 of 63 MAFALDA and 0 of 60 Argotario: 3 of 183 overall---rare in every negative class, whatever the protocol. We give raw counts because the samples are small; ``3\%'' is one item. Deliberate search finds them no more easily: of 25 hand-selected candidates, the same annotators confirmed 10 (\S\ref{sec:natural}).

Non-argument rate does not track the native false-positive rate. It runs 3\% on CoCoLoFa, 40\% on Reddit, 52\% on MAFALDA and 65\% on Argotario, and the corpus with the least, CoCoLoFa, has the highest native rate. CoCoLoFa's class is instead almost entirely genuine argument (93\% category B, one non-argument in 30), consistent with its role as the hardest case (\S\ref{sec:setup}). MAFALDA's low native rate has a different source: its taxonomy is partly disjoint, so almost none of its valid class instantiates a scheme inside CoCoLoFa's label space.

\subsection{Scheme recognition scored as fallacy detection}
\label{sec:central}

Table~\ref{tab:fpr} reports false-positive rates on the native valid class and on matched negatives, across four configurations. Matched negatives are over-flagged far above the native rate in every one; seed-to-seed variation is small relative to the gaps; the effect is not architecture-specific.\footnote{Single-seed CoCoLoFa run: matched 57.3\%, native 13.6\%, with non-overlapping bootstrap 95\% CIs [53.6\%, 61.1\%] and [9.8\%, 17.4\%].} The wrong-scheme column runs higher still (93.5\% on CoCoLoFa): those are flag rates over a condition certified valid at 77.5\% (\S\ref{sec:wsvalidity}), and the informative quantity is not that level but which type the flags assign.

\begin{table*}[t]
\centering
\small
\setlength{\tabcolsep}{5pt}
\begin{tabular}{lccccc}
\toprule
Model & Native & Matched & WS (raw)$^{\dagger}$ & Type-specific & N$\rightarrow$M \\
\midrule
CoCoLoFa / ModernBERT (5 seeds) & 16.6\%$\pm$1.6\% & 58.9\%$\pm$2.9\% & 93.5\%$\pm$1.8\% & $+40.9\pm1.9$pp & 42.3pp \\
CoCoLoFa / RoBERTa (3 seeds) & 19.7\%$\pm$3.3\% & 70.3\%$\pm$3.2\% & 96.5\%$\pm$0.8\% & $+48.5\pm3.9$pp & 50.7pp \\
Reddit / ModernBERT, within-corpus (3) & 5.7\%$\pm$0.1\% & 62.0\%$\pm$2.8\% & 60.2\%$\pm$4.0\% & $+32.9\pm3.4$pp & 56.3pp \\
Reddit, cross-corpus & 16.3\% & 42.9\% & n.r. & n.r. & 26.6pp \\
\bottomrule
\end{tabular}
\caption{\label{tab:fpr} False-positive rates (means$\pm$SD across seeds). WS = wrong-scheme; type-specific = the gap in how often each condition is classified as the source type $\tau$; N$\rightarrow$M = native-to-matched gap; \emph{n.r.} = not computed. $^{\dagger}$Wrong-scheme columns are flag rates, not false-positive rates: the condition is certified valid at 77.5\% (\S\ref{sec:wsvalidity}), so some fraction of those flags is correct.}
\end{table*}

\paragraph{The effect is specific to the scheme.} Averaged over the eight CoCoLoFa types, a matched negative is classified as its source type $40.9 \pm 1.9$ points more often than an equally careful wrong-scheme negative. For seven of the eight the wrong-scheme rate is zero; per-type gaps run from $+10.7$ to $+73.1$pp, and no per-type matched-rate interval overlaps its wrong-scheme interval. A wrong-scheme negative built against $\tau$ is itself a correct instance of some other scheme $\tau'$, and the classifier finds it. The classifier labels them with the scheme they instantiate 85.9\% of the time, and with the source type 0.4\% of the time---a more than 200-to-1 preference. Per type, $\tau'$ ranges from 76.3\% for false dilemma to 94.0\% for appeal to nature, and $\tau$ never exceeds 2.2\%. Because the conditions differ only in scheme identity, this is a claim about what the classifier tracks, not about how hard the items are: it has learned which scheme an argument uses, not whether it uses it correctly. On the benchmark's own test set these two abilities are indistinguishable: every item instantiating a tracked scheme is labelled with a fallacy, so a model that recognizes the scheme scores as a model that detects the fallacy. What the benchmark reports as fallacy detection is scheme recognition, and it contains no items on which the two would come apart. An item-level logistic regression over the 1{,}393 constructed items finds condition dominant and readability with no independent effect ($p = 0.86$); with only three of 738 wrong-scheme items classified as the source type, the Firth-penalized odds ratio is 144 (penalized likelihood-ratio $p < 0.001$). Applied to a deliberately built negative class the same measurement yields a far lower rate (\S\ref{sec:touche}), so this is a property of the class, not of the procedure.

\paragraph{Domain shift does not explain it.} A CoCoLoFa-trained model evaluated on Reddit invites a domain-shift reading, so we removed the domain gap by training on Reddit from scratch. Domain shift predicts the gap should shrink; it grows, from 26.6pp to 56.3pp. Raw growth alone is not decisive, since a more sensitive model flags more of everything; that is also why this model's raw wrong-scheme rate is elevated. The within-corpus model still classifies matched negatives as their own source type $32.9 \pm 3.4$ points more often than wrong-scheme negatives, with five of eight types at zero, and domain shift produces no such asymmetry.

\paragraph{Corroborating corpora.} Of Argotario's two types inside CoCoLoFa's label space, Irrelevant Authority shows the effect strongly: matched 78.8\% against native 11.7\%. Hasty generalization does not, at 6.8\%, below the native rate. Its matched negatives also had the lowest judge precision of any type in the validation study (66.7\%, \S\ref{sec:human}). MAFALDA's qualitative check is consistent with the effect (Table~\ref{tab:corrob}).

\paragraph{Two types behave differently.} Six types show type-specific gaps between $+31.8$ and $+73.1$pp. Appeal to nature ($+10.7$pp single-seed, $+7.9$ multi-seed) and false dilemma ($+19.4$, $+23.8$) are weaker. False dilemma is a structural outlier on Walton's account, and is also the weakest type on Reddit ($+2.1$pp). Appeal to nature is the weakest type on every detector we tested, encoder and zero-shot alike, which is consistent with the surface term ``natural'' acting as a lexical cue (see Limitations).

\subsection{The effect is not specific to fine-tuned encoders}
\label{sec:llm}

Encoders fine-tuned on a diluted negative class could have learned the dissociation from it. We repeated the measurement with three zero-shot detectors on the identical items: Claude Sonnet 5 and Haiku 4.5, varying scale within one family, and Qwen3.6-27B, varying training lineage. All ran without chain-of-thought, with the prompt frozen before the first call (Appendix~\ref{app:prompts}); 2 of 10{,}260 responses failed to parse.

It reproduces on all three (Table~\ref{tab:transfer}). Wrong-scheme items are labelled with the scheme they instantiate 69.9--86.3 points more often than with the source type, and the source-type rate stays between 0.3\% and 0.5\% against the encoders' 0.4\%. Haiku 4.5 lands closest to the encoders on both arms: a type-specific gap of $+45.6$ against their $+40.9$, and a $\tau'$-over-$\tau$ margin of $+86.3$ against their $+85.5$ (the 85.9\% and 0.4\% of \S\ref{sec:central}), while Sonnet 5, the most capable detector tested, shows the effect most weakly of the Claude pair and still separates the conditions by $+31.0$pp. The effect attenuates with capability rather than appearing with it, and no detector escapes it.

\paragraph{The raw gap does not survive; the dissociation does.} The native-to-matched gap is $+13.5$pp for Sonnet 5 and $+12.6$ for Haiku 4.5, but $+0.5$pp for Qwen3.6-27B, whose interval spans zero: on that detector the raw measure shows no effect at all. The same items give it a type-specific separation of $+25.5$pp and a $\tau'$ margin of $+69.9$pp. On both Claude models the gap reverses sign under a binary prompt posing only the detection job (Sonnet 5 $-7.7$pp, Haiku 4.5 $-33.2$pp), all four intervals excluding zero; under the same prompt Qwen3.6-27B's null typed gap turns clearly negative ($-25.6$pp). Naming a type forces a commitment that native items, instantiating no tracked scheme, mostly fail. The rate is therefore unstable across prompts and across models, while the comparison between two conditions built by one pipeline is stable across both.

\begin{table}[t]
\centering
\small
\setlength{\tabcolsep}{3pt}
\begin{tabular}{lcccc}
\toprule
Detector & Native & Matched & WS$^{\dagger}$ & M$-$WS as $\tau$ \\
\midrule
ModernBERT ft. & 16.6\% & 58.9\% & 93.5\% & $+40.9\pm1.9$ \\
Sonnet 5 & 24.1\% & 37.6\% & 74.1\% & $+31.0$ \\
Haiku 4.5 & 53.9\% & 66.6\% & 90.4\% & $+45.6$ \\
Qwen3.6-27B & 32.5\% & 33.0\% & n.r. & $+25.5$ \\
\midrule
\multicolumn{5}{l}{\emph{deliberately built class (Reddit-derived)}} \\
ModernBERT ft. & 16.3\% & 42.9\% & \multicolumn{2}{l}{Touch\'e: 26.2\%} \\
\bottomrule
\end{tabular}
\caption{\label{tab:transfer} The measurement across detectors, typed prompt, CoCoLoFa unless noted. M$-$WS as $\tau$ is the type-specific separation in percentage points. Qwen3.6-27B shows no native-to-matched gap ($+0.5$pp, interval spanning zero) yet a $+25.5$pp separation on the same items. $^{\dagger}$Wrong-scheme columns are flag rates, not false-positive rates (\S\ref{sec:wsvalidity}). Binary-prompt rates are in the text. Last row: the cross-corpus baseline and our matched negatives against Touch\'e's negative class (\S\ref{sec:touche}).}
\end{table}

\subsection{A negative class built deliberately}
\label{sec:touche}

Every benchmark in \S\ref{sec:composition} built its valid class without scheme-level targeting, so the audit has only been applied where it was expected to find a gap. The Touch\'e 2026 shared task \citep{touche2026} supplies the missing case: its non-fallacious class was assembled by selecting arguments that resemble a specific fallacy type.

The label spaces coincide: all 473 non-fallacious entries carry a resemblance tag, and all eight tags map onto CoCoLoFa's eight types, 55--64 items each. The corpus is Reddit-derived, so the baseline is the cross-corpus row of Table~\ref{tab:fpr}: the CoCoLoFa-trained model flags Reddit's native class at 16.3\%, where 5.7\% elsewhere is the within-corpus model on the same class; rates are in Table~\ref{tab:transfer}. The audit discriminates, coming in well below the 42.9\% on our matched negatives. And resemblance-selection recovers roughly a third of the distance that critical-question matching recovers: selecting arguments that resemble a fallacy is not equivalent to constructing arguments that answer the scheme's critical question. When flagged, Touch\'e items are called the type they resemble only $15.1 \pm 1.7\%$ of the time.

\paragraph{Register does not explain the gap.} Our matched negatives are machine-written and more polished than raw forum text, so classifiers might flag them for register. Touch\'e bounds this directly: it contains the same arguments raw and in a self-contained rewrite folding in the parent comment. The rewrite is 9.2 words longer on average and 23 points higher in contraction rate, and it moves the false-positive rate from $24.6 \pm 1.9\%$ to $26.2 \pm 1.8\%$: 1.6 points, smaller than the between-seed standard deviation. A register shift of that size cannot produce a 16.7-point gap. The raw items are also register-matched to our native Reddit class (46.5 against 42.7 words, 52.2\% against 51.0\% contractions).

A further Touch\'e version rewritten with knowledge of the fallacy and scheme labels shows $36.6 \pm 1.2\%$ on identical arguments; we exclude it as a second construction artifact in the same benchmark.

\subsection{The effect appears on text no one wrote for this study}
\label{sec:natural}

Every result above uses generated negatives. As a check that depends on none of them, we identified scheme-matched valid arguments already present in the native ``none'' classes: 25 hand-selected candidates from CoCoLoFa, Reddit and Argotario, pooled with the \S\ref{sec:composition} samples and coded by the same two annotators, blind to which items were targeted. The annotators confirmed 10 of the 25 candidates, which with the 3 found in the random samples gives 13 confirmed scheme-matched arguments. Deliberate search barely finds such arguments; the pool is not a base-rate sample and does not revise \S\ref{sec:composition}.

The classifier flags 7 of the 13 (53.8\%; 95\% CI [29.1\%, 76.8\%]), against 19.1\% of the genuine arguments instantiating no tracked scheme (21 of 110) and 5.9\% of non-arguments (5 of 85). That comparison is the wrong-scheme contrast occurring naturally: A and B are both genuine arguments, differing only in whether a tracked scheme is present. Flag rate rises with scheme content, on text written years before this study. The scheme-specificity is sharp as well. Of the six flagged items with a pinned target scheme, five are classified as exactly that scheme; the sixth, a citation of scientific consensus on climate change, is labelled appeal to majority---a confusion between neighbouring schemes, not a generic over-flag.

\subsection{Human annotators confirm the negatives are valid}
\label{sec:human}

The reported rate would be inflated if the judge had admitted subtle fallacies, so we checked it. Three independent annotators, none of them authors, judged 72 matched negatives, 56 from CoCoLoFa and 16 from Reddit, as Valid, Invalid or Borderline under rules fixed in advance (protocol in Appendix~\ref{app:annotation}). The items came from the judge's \emph{retained} set, the one the headline rate is computed over; annotators saw only the argument text with its target scheme and critical question, never classifier output. Eight benchmark-labelled fallacies were planted and excluded from all figures; annotators caught seven, seven and six of them.

The judge's precision on its retained set is 93.1\% (67 of 72): 94.6\% on CoCoLoFa (53 of 56), 87.5\% on Reddit (14 of 16). Cohen's $\kappa$ is deflated by the prevalence of Valid items, so we report Gwet's AC1 at 0.69 binary and 0.74 three-way.

Restricted to the human-confirmed-valid CoCoLoFa subset, the false-positive rate is 56.6\% (30 of 53; 95\% CI [43.4\%, 69.8\%]) against a native 16.6\%, a gap of 40.0 points, of which judge impurity accounts for roughly one point. The classifier over-flags more than half of the arguments three blind annotators confirmed valid. Reddit agrees (57.1\%, 8 of 14, against 5.7\%) on a sample too small to carry the claim.

\subsection{Validity of the wrong-scheme condition}
\label{sec:wsvalidity}

The judge never checked critical-question satisfaction for wrong-scheme items, so the raw rates in Table~\ref{tab:fpr} sit on an uncertified population. A separate, non-overlapping panel repeated the \S\ref{sec:human} protocol against each item's target scheme ($n{=}40$; Table~\ref{tab:human}): precision is 77.5\% (95\% CI [62.5\%, 87.7\%]) against 93.1\% for matched items, the shortfall concentrated in the appeal-to-nature and false-dilemma target schemes. The type-specific statistic is unaffected: a genuine fallacy of its own scheme is still not a fallacy of the source type.

\section{Conclusion}
\label{sec:conclusion}

Across four benchmarks, scheme-matched valid arguments make up a few percent at most of the valid class, so the reported rate averages over the wrong population. On the right one---wrong-scheme arguments labelled with their own scheme 85.9\% of the time, with the scored type 0.4\%---these classifiers have learned which scheme an argument uses, not whether it uses it correctly. A reported false-positive rate should not be trusted as a measure of detection until the valid class has been audited for scheme-matched coverage.

\section*{Limitations}
\label{sec:limitations}

Our generated rates are not estimates of deployment-time magnitude. A rate measured on constructed items is a joint function of classifier and generator, so the 58.9\% and 62.0\% figures should not be read as the rate a deployed detector would show, and the naturally-occurring check (\S\ref{sec:natural}) is too small (13 confirmed items) to pin that magnitude on its own.

The \S\ref{sec:composition} samples remain small (183 items, three of them category A), so the percentages are a direction, not a prevalence. The matched negatives carry a 6.9\% impurity. The validation sample is modest ($n{=}72$, with a 16-item Reddit subset), the judge is unvalidated on Argotario and MAFALDA, and the two validation panels were different groups on a task where judgment varies. The wrong-scheme condition is certified valid at only 77.5\% (\S\ref{sec:wsvalidity}).

The zero-shot check (\S\ref{sec:llm}) covers three detectors on CoCoLoFa only, and the only open-weights detector tested is mid-size. Wrong-scheme flag rates throughout are flag rates, not false-positive rates: 22.5\% of that condition is not certified valid (\S\ref{sec:wsvalidity}), so some fraction of those flags is correct. The Touch\'e comparison (\S\ref{sec:touche}) is between different items, not counterparts of ours, so construction protocol is not the only thing that differs between the two negative classes; the register analysis bounds one alternative and not the others. Argotario's corroboration is partial: of its two types inside CoCoLoFa's label space, one shows the effect and one does not. Argotario and MAFALDA enter the rarity finding but not the primary quantitative claims.

We use Walton's schemes instrumentally; their individuation has been challenged \citep{katzav2004classification}, a dispute our argument does not need to resolve, since it requires only that some theory-grounded partition of ``same reasoning move'' works for the schemes a taxonomy already uses. We do not operationalize pragma-dialectics.

The constructed negatives are not used to replace any benchmark's valid class; building and validating a corrected benchmark, and showing that it improves detection, is a separate contribution that presupposes the measurement problem documented here. Scheme Foils covers eight fallacy types on two corpora, and the audit's coverage is bounded by the schemes a taxonomy already tracks; we open the released sets for extension to further schemes and corpora. Whether the size of the gap tracks a scheme's prevalence among natural valid arguments, and whether equalizing per-scheme representation lowers false-positive rates, are untested.

\section*{Ethical Considerations}

The annotators in all three studies (\S\ref{sec:composition}, \S\ref{sec:human}, \S\ref{sec:wsvalidity}) were volunteers who consented to participate and were not compensated; they were briefed on the task and were not exposed to harmful content beyond ordinary online argumentation. Generation prompts instruct the model to substitute a related defensible point where a source item rests on a harmful, discriminatory or dehumanizing premise. All source corpora are publicly released research datasets and are used within their licences; we redistribute source items only where licensing permits.

\bibliography{refs}

\appendix

\section{Training Details}
\label{app:training}

The CoCoLoFa ModernBERT models reached $81.4\% \pm 0.5\%$ dev accuracy across five seeds and the RoBERTa models $81.9\% \pm 0.8\%$, against a 39.7\% majority baseline; the Reddit models reached $60.4\% \pm 0.2\%$ across three seeds against a 49.0\% baseline. The Reddit models trained six epochs against three, since token-span label aggregation produces a noisier signal, under an identical selection criterion. All classifiers were fine-tuned with AdamW (weight decay 0.01), a linear schedule with 10\% warmup, learning rate 2e-5, batch size 16, maximum sequence length 256, gradient clipping at 1.0, and no gradient accumulation. The multi-seed models trained on a single NVIDIA T4; the single-seed run of footnote 1 and all inference ran on a 6\,GB RTX 4050 laptop GPU.

\paragraph{Corpus details.} CoCoLoFa labels news-article comments with eight fallacy types---appeal to authority, appeal to majority, appeal to nature, appeal to tradition, appeal to worse problems, false dilemma, hasty generalization, slippery slope---plus a ``none'' class. The Reddit corpus's eight-class annotation is mapped onto CoCoLoFa's scheme (its ``Black-or-white'' is false dilemma). MAFALDA is expert-annotated over a partly disjoint taxonomy.

\paragraph{Statistical procedures.} The permutation test shuffles the model's predicted labels while preserving its overall rate of predicting ``none.'' In the item-level logistic regression the two conditions are near-perfectly separated---only three of 738 wrong-scheme items are classified as the source type---so the unpenalized odds ratio is unstable and the Firth-penalized refit is reported.

\section{Annotation Protocols}
\label{app:annotation}

\paragraph{Matched negatives (\S\ref{sec:human}).} Three independent undergraduate volunteers, not authors, briefed and calibrated, told only the coding task and not what the study was testing, consenting and unpaid. The CoCoLoFa sample was stratified seven per type across the eight types. Judgments were three-way---Valid, Invalid, Borderline---under rules fixed in advance: majority vote decides, Borderline never counts as Valid, ties break conservatively. Per-type judge precision was 100\% for six of the eight types, 77.8\% for appeal to authority and 66.7\% for hasty generalization.

\paragraph{Wrong-scheme negatives (\S\ref{sec:wsvalidity}).} A separate panel of three annotators of the same background, given the same briefing and calibration and non-overlapping with the first panel, judged each item ($n{=}40$) against the target scheme it was built to instantiate, catching six, four and six of six planted decoys.

\paragraph{None-class characterization (\S\ref{sec:composition}, \S\ref{sec:natural}).} Two independent annotators of the same background, overlapping with neither panel, coded all 208 items---the four corpus samples and the candidate pool---blind to which items were hand-selected, to all prior codes, and to classifier output. A joint calibration pass on 10 items drawn from outside the coded pool preceded independent coding. Disagreements were resolved by discussion without the authors, under a rule fixed in advance that unresolved items default to the non-A code; none required the default. Six benchmark-labelled fallacies were planted and excluded from all figures; the annotators caught four and five of them.

\paragraph{Scheme Foils.} We release the judge-filtered sets under the name Scheme Foils: 655 matched and 738 wrong-scheme CoCoLoFa items, 387 matched and 389 wrong-scheme Reddit items, with target schemes, critical questions, and source items where licensing permits. Each item carries the human-validated precision estimate of the study it belongs to, so a user knows what impurity they inherit (6.9\% for the matched split, 22.5\% for the wrong-scheme split). The intended use is direct: score an already-trained detector on the matched split against its own reported false-positive rate, and use the wrong-scheme split to check that any gap is scheme-specific rather than generic. Neither requires retraining or regeneration. The sets are available at \url{https://github.com/fine2006/the-concealment-hypothesis}.

\section{Generation Prompts}
\label{app:prompts}

Per-type prompts were calibrated to each corpus's register, one prompt design per fallacy type, stating the target scheme and its critical question explicitly. The slots \texttt{cq\_block}, \texttt{target\_cq\_block}, \texttt{source\_type}, \texttt{target\_type} and \texttt{examples\_block} are filled per batch. Batches of 15--20 items are generated per call, with strict numbered-output parsing and up to two retries on misalignment. Automated first-pass checks flag source-scheme keyword leakage in wrong-scheme outputs and compare word-count, sentence-count and formal-register statistics between conditions.

\textit{Scheme and critical-question blocks (one per fallacy type; abridged to three representative types):}
\begin{lstlisting}
[appeal to authority]
SCHEME: Expert Opinion. The fallacy cites a source who is NOT a genuine domain expert for the claim being made.
A SATISFYING FIX replaces the source with a genuinely relevant domain expert -- add ONLY the minimum credential needed, nothing more (no years of experience, no institutional pedigree).

[hasty generalization]
SCHEME: Generalization from Sample. The fallacy draws a broad conclusion from a sample too small or unrepresentative to support it (often a personal anecdote).
A SATISFYING FIX replaces the thin/anecdotal sample with one that is genuinely adequate -- a real, larger, more systematic basis for the same conclusion. Do NOT just hedge the conclusion ("might," "could") -- ground it instead.

[false dilemma]
SCHEME: Disjunctive reasoning (closest to formal logic, not a defeasible Waltonian scheme -- the hardest type to fix cleanly).
A SATISFYING FIX shows the binary is genuinely exhaustive in THIS specific context -- e.g. cite why middle options have specifically failed or been closed off here -- OR explicitly names a real third option that resolves the false binary.
\end{lstlisting}

\textit{Matched-negative prompt (CoCoLoFa):}
\begin{lstlisting}
You will be given a list of numbered fallacious arguments, all of the SAME fallacy type. Your task is to write a FRESH, standalone valid argument for each one -- not an edit of the original -- written exactly the way an ordinary person commenting online would write it.

{cq_block}

THE TASK:
For each example, write a new valid argument on the SAME topic, in the SAME casual register, with the SAME approximate length and sentence count as the original (most originals are 2-5 sentences). Do not copy or echo specific distinctive phrases from the original.

CRITICAL -- VARY HOW YOU SIGNAL GROUNDING:
Do not use the same template repeatedly across this batch. Mix it up: sometimes a concrete specific fact stated plainly, sometimes reported speech from a named type of person, sometimes a first-person framing, sometimes a plain declarative claim with no explicit evidence-marker.

THE RULE THAT DECIDES CONFLICTS:
The fix must be genuinely correct -- but correctness should be expressed at the SAME natural, unforced confidence level an ordinary person would use, not "proven" through extra explanation or citation. The bar is: a reasonable person would NOT consider it an error to call it valid.

DO NOT CHANGE WHAT IS BEING ARGUED FOR:
The fix must support the SAME conclusion as the original -- only the reasoning changes.

WHAT TO AVOID:
- Citation-dense outputs ("Studies show...") -- this reads as an AI-generated abstract.
- Adding credentials or statistics beyond what the fix needs.
- Tacking on an unverified conditional as the only grounding.
- If the claim rests on a religious/contested-worldview premise, write "SKIP -- non-empirical premise".

OUTPUT FORMAT: a numbered list, one entry per input number, in order, with NOTHING else.
[number]. [your fresh valid version]

Here are the examples:
{examples_block}
\end{lstlisting}

\textit{Wrong-scheme prompt (CoCoLoFa):}
\begin{lstlisting}
You will be given numbered fallacious arguments, all of the SAME fallacy type ({source_type}). Your task is NOT to fix that fallacy. Instead, write a FRESH, valid argument for each that stays on the SAME topic, with the SAME effort, length and casual register a normal fix would have -- but uses a COMPLETELY DIFFERENT reasoning style: {target_type}.

THE SCHEME YOU MUST ACTUALLY USE ({target_type}):
{target_cq_block}

CRITICAL -- DO NOT USE {source_type}'S REASONING AT ALL:
Completely avoid {source_type}'s characteristic content and language. Stay on the same topic, but make the argument using {target_type}'s logic, as if you were never thinking about {source_type}.

NOTE ON CONCLUSION: unlike a normal fix, this does NOT need to argue for the same specific conclusion. Staying on the same general topic with comparable effort is what matters.

SELF-CHECK: if someone read your answer without knowing the target scheme, would they identify it as {target_type}? If it could still be read as {source_type}, you have not switched schemes.

{base_rules}

OUTPUT FORMAT: a numbered list, one entry per input number, in order, with NOTHING else.
[number]. [your fresh {target_type} argument, same topic as the source]

Here are the examples:
{examples_block}
\end{lstlisting}

\textit{Zero-shot detector prompt (\S\ref{sec:llm}); frozen before the run:}
\begin{lstlisting}
Classify the following argument. If it commits a logical fallacy, answer with the fallacy type from this list:
- appeal to authority
- appeal to majority
- appeal to nature
- appeal to tradition
- appeal to worse problems
- false dilemma
- hasty generalization
- slippery slope

If it does not commit a fallacy, answer with: none

Answer with the label only, nothing else.

Argument: {text}

Answer:
\end{lstlisting}

The binary variant asks only: \texttt{Does the following argument commit a logical fallacy? Answer with one word: yes or no.} All detectors ran zero-shot with thinking disabled and a 64-token cap. Both prompts ran on all three detectors, 10{,}260 responses in total, of which 2 failed to parse. Model identifiers: \texttt{claude-sonnet-5} and \texttt{claude-haiku-4-5} via the Anthropic API and \texttt{qwen3.6-27b}, all queried 18--20 August 2026; the generator throughout is \texttt{gemini-2.5-flash} and the judge \texttt{deepseek-v3}.

The Reddit variants of both prompts are identical in structure, with the source item given as a SPAN plus its parent comment as CONTEXT, instructions to match the shorter Reddit register, and a SKIP option for spans too short or context-dependent to work with.

\section{Complementary Tables}
\label{app:tables}

\begin{table}[!ht]
\centering
\small
\setlength{\tabcolsep}{4pt}
\begin{tabular}{lcccc}
\toprule
Cat. & CoCo (30) & Redd (30) & MAF (63) & Argo (60) \\
\midrule
A & 1 (3\%) & 1 (3\%) & 1 (2\%) & 0 (0\%) \\
B & 28 (93\%) & 17 (57\%) & 29 (46\%) & 21 (35\%) \\
Non-arg. & 1 (3\%) & 12 (40\%) & 33 (52\%) & 39 (65\%) \\
\bottomrule
\end{tabular}
\caption{\label{tab:composition} Composition of the native ``none'' class across four corpora by two independent annotators (sample size in parentheses; MAFALDA's extractable none class coded in full). A = scheme-matched valid argument; B = genuine argument, no tracked scheme; Non-arg.\ = not an argument (fragments, questions, noise, off-topic remarks; two items are meta-commentary about fallacies).}
\end{table}

\begin{table}[!ht]
\centering
\small
\setlength{\tabcolsep}{3pt}
\begin{tabular}{lccc}
\toprule
Study & $n$ & Precision & AC1 \\
\midrule
Matched (\S\ref{sec:human}) & 72 & 93.1\% (67/72) & 0.69 / 0.74 \\
Wrong-scheme (\S\ref{sec:wsvalidity}) & 40 & 77.5\% [62.5, 87.7] & 0.81 \\
\bottomrule
\end{tabular}
\caption{\label{tab:human} Human validation of both constructed conditions. Precision is the proportion of the judge's retained set confirmed valid by majority vote; planted decoys are excluded from all precision figures. AC1 is Gwet's, binary / three-way for the matched study. Panels were separate and non-overlapping.}
\end{table}

\begin{table}[!ht]
\centering
\small
\setlength{\tabcolsep}{2pt}
\begin{tabular}{lc}
\toprule
Quantity (CoCoLoFa-trained ModernBERT) & Value \\
\midrule
Argotario native FPR & 11.7\% \\
Argotario matched FPR (all five types) & 16.4\% \\
Argotario matched, Irrelevant Authority & 78.8\% \\
\quad raw native$\rightarrow$matched gap & 67.1pp \\
\quad type-specific gap & $+72.7$pp \\
Argotario matched, hasty generalization & 6.8\% (3/44) \\
\midrule
MAFALDA native FPR & 7.9\% \\
MAFALDA overlapping fallacy spans flagged & 50.5\% \\
\quad identified by exact type & 41\% \\
\bottomrule
\end{tabular}
\caption{\label{tab:corrob} Corroborating-corpus results (\S\ref{sec:central}). Three of Argotario's five types (ad hominem, appeal to emotion, red herring) fall outside CoCoLoFa's label space and are uninformative by construction; the two that overlap are reported separately. MAFALDA is evaluated qualitatively, its native rate low for the label-space reason given in \S\ref{sec:composition}.}
\end{table}

\end{document}